# Co-occurrence of the Benford-like and Zipf Laws Arising from the Texts Representing Human and Artificial Languages

Evgeny Shulzinger[a], Irina Legchenkova[b] and Edward Bormashenko[b]

[a]Ariel University, Faculty of Natural Sciences, Department of Physics, Ariel, P.O.B. 3, 407000, Israel

[b]Ariel University, Faculty of Engineering Sciences, Department of Chemical Engineering Biotechnology and Materials, Ariel, P.O.B. 3, 407000, Israel

Corresponding author: Edward Bormashenko, Ariel University, Faculty of Engineering Sciences, Department of Chemical Engineering Biotechnology and Materials, Ariel, P.O.B. 3, 407000, Israel. Phone: 972-0747296863; Fax: 972-39066621;

E-mail: edward@ariel.ac.il

**Abstract**

We demonstrate that large texts, representing human (English, Russian, Ukrainian) and artificial (C++, Java) languages, display quantitative patterns characterized by the Benford-like and Zipf laws. The frequency of a word following the Zipf law is inversely proportional to its rank, whereas the total numbers of a certain word appearing in the text generate the uneven Benford-like distribution of leading numbers. Excluding the most popular words essentially improves the correlation of actual textual data with the Zipfian distribution, whereas the Benford distribution of leading numbers (arising from the overall amount of a certain word) is insensitive to the same elimination procedure. The calculated values of the moduli of slopes of double logarithmical plots for artificial languages (C++, Java) are markedly larger than those for human ones.

**1. Introduction**

Human languages demonstrate hidden phonetic, semantic and historical patterns, which may be revealed with computer analysis (Bybee, J. L. & Hopper, 2001; Johnson, 2008, Gerlach & Altmann, 2013; Aiden & Michel, 2013; Shulzinger & Bormashenko, 2017). The origin of these patterns remains mysterious; however, counter to intuition, significant pattern-generating evolution of linguistic behavior was observed by computer simulations (Kirby, 2001). From an initially unstructured communication system (a protolanguage) a fully compositional syntactic meaning-string mapping emerged (Kirby, 2001).

One of remarkable features of natural languages is Zipf's law (Zipf, 1965a; Pustet, 2004). Zipf's law states that in all natural languages, the frequency of a word is inversely proportional to its rank (Zipf, 1965, Manin, 2009, Fontanari & Perlovsky, 2004, Fontanari & Perlovsky, 2009, Mehri & Lashkari, 2016), namely Eq. 1 takes place:

$$f(n) = \frac{C}{n^\alpha}, \quad (1)$$

where $C$ is the constant, $\alpha \cong 1$. Zipf's distribution was observed for languages as diverse as English, Latin, German, Gothic, Chinese (Mandarin), Lakota, Nootka, and Plains Cree (Pustet, 2004). Baek et al. draw attention to the fact that the value of $\alpha$ in not necessarily close to unity; and the general power-law supplied by Eq. 1 is expected for languages (Baek et al., 2011).

Zipf suggested that the revealed frequency distribution arises from the "least effort" principle requested from transport of information from speaker to listener (Zipf 1965b). This hypothesis is supported by recent investigations (Ferrer i Cancho & Solé, 2003, Ferrer i Cancho, 2005). It was also suggested that the recurrence of Zipf's law in human languages could originate from pressure for easy and fast communication (Ferrer-i-Cancho 2016).On the contrary, it was suggested that Zipf' distribution arises from the Poisson process (Eliazar 2016; Reed 2001).

However, both the roots and significance of Zipf's law in linguistics remain unclear (Fontanari & Perlovsky, 2004). On the one hand, the finding that texts produced by the random emission of symbols and spaces, so that words of the same length are equiprobable, also generate word frequency distributions that follow a generalized Zipf's law, also called the Zipf-Mandelbrot law (Mandelbrot, 1982; Manin, 2009). It was also noted that Zipf's law is not valid for the most common words (Tsonis et al., 1997a). Montemurro noted that the Zipf–Mandelbrot law can only describe the statistical behavior of a rather restricted fraction of the total number of words contained in some given corpus (Montemurro, 2001). Significant deviations between the predicted hyperbolic and real frequency of different words of a text were reported (Ferrer-i Cancho & Solé, 2001). Moreover, it was demonstrated that various random sequences of characters (random texts) reproduce Zipf's law (Ferrer-i-Cancho & Elvevåg 2010).

On the other hand, quantitative analyses of the evolution of syntactic communication in comparison to animal communication (Nowak et al., 2000) and the emergence of irregularities in language (Kirby, 2001) latently assume that human lexicons follow the Zipf distribution of word frequencies (Fontanari & Perlovsky, 2009). Zipf himself related structuring of languages to a ''principle of least effort'', resembling the principle of least action in physics (Zipf, 1965b, p. 20; Lanczos, 1970). It is also noteworthy that Zipf-like distribution was revealed in a plenty of statistical problems, including the distribution of firm sizes (Stanley et al., 1995), distributions of open spaces in city space (Volchenkov & Blanchard 2008) and genomic data (Mantegna et al., 1994; Tsonis & Tsonis, 2002). Mandelbrot suggested that Zipf's hyperbolic frequency distributions are about as prevalent in social sciences as Gaussian distributions are dominating in the natural sciences (Mandelbrot, 1982). It was also demonstrated that in certain cases, Zipf ranking appears as the ordering by growing Kolmogorov complexity (Manin, 2014).

As to the roots of the origin of the Zipf law, Baek et al., suggested that Zipf's law arises from the random group formation phenomena (Baek et al., 2011). The elements of a group can be citizens of a country, or the groups of family names, or the elements can be all the words making up a novel, or the groups comprised of unique words. Thus, it was suggested that the Zipf law has a pure statistical origin (Baek et al., 2011). It was also conjectured that Zipf's law is related to the focal expression of a generalized thermodynamic structure (Altamirano & Robledo, 2011). This structure is obtained from a deformed type of statistical mechanics that arises when configurational phase space is incompletely visited in a strict way; the restriction is that the accessible fraction of this space has fractal properties (Altamirano & Robledo, 2011). It was also assumed that Zipf's law arises from the fractal structure of the language itself (Mehri & Lashkari, 2016). It was conjectured that Zipf's distribution may be the ultimate signature of an integrated ("coherent") system (Cristelli et al., 2012).

Thus we conclude that the origin of Zipf distribution remains at least debatable (Pustet, 2004); Gell-Mann stated that ''Zipf's law remains essentially unexplained'' (Gell-Mann, 1994, p. 97). It is reasonable to suggest, that understanding the origin of the Zipf distribution will help to answer the question: what is the precise meaning of the notion of "language" and what are characteristic features of languages? Do DNA codes and programming languages exemplify true languages (Tsonis, 1997b)? It was demonstrated recently that the Zipf-like distribution appears in computer programs (Zhang, 2009).

An intimate relative of the Zipf Law is the Benford law, originally discovered in 1881 by Simon Newcomb. Newcomb noticed the surprising pattern of wear of pages of the logarithm tables. He established that the earlier pages of the tables were used much more than the later pages. The earlier pages contained numbers starting from the character "1", leading Newcomb to suggest that the distribution of leading figures is unequal (Newcomb, 1881). He proposed the semi-empirical logarithmic law describing the occurrence of first significant digits (also called the leading digits) in statistical data:

$$P(d) = \log_{10}\left(1 + \frac{1}{d}\right), \qquad d = 1, 2, ..., 9, \qquad (2)$$

where $P(d)$ is the probability of a number having the first non-zero digit $d$.

The logarithmical distribution was re-discovered by Frank Benford (Benford, 1938), and since then it is labeled "the Benford distribution". The Benford distribution of leading digits was registered in a broad diversity of statistical data (Benford, 1938; Mir, 2012; Li et al., 2015; Karthik et al., 2016). Researchers have already drawn attention to the deep relationship between the Newcomb-Benford distribution of leading digits (Benford, 1938; Mir, 2012; Sambridge, 2010;

Bormashenko et al., 2016; Li & Fu, 201; Friar, Goldman, & Pérez-Mercader, 2016; Barabesi et al., 2017) and the Zipf law (Pietronero et al., 2001; Pustet, 2004; Baek et al., 2011).

Pietronero et al. related the origin of these laws to a broad ubiquity in natural systems, demonstrating scaling invariance (Pietronero et al., 2001). The roots of both of them were related to the scaling invariance of complex objects (Pietronero et al., 2001), thermodynamic properties of statistical systems (Altamirano & Robledo, 2011), the principle of least effort (Pustet, 2014, Ferrer i Cancho & Solé, 2003, Ferrer i Cancho, 2005) and properties of the place-to-value notation accepted for representing various sets of data (Whyman, & Bormashenko, 2016a).

We demonstrate that texts representing various human languages and programs written in Java and C++ generate both Zipfian (with various values of $α$ appearing in Eq. 1) and Benford distributions, where the frequencies of words follow the Zipf law, whereas total numbers of a certain word give rise to the Benford-like distribution of leading numbers.

## 2. Methods

The treated texts were: in English (Joanne K. Rowling, "*Harry Potter and the Chamber of Secrets*", James Joyce "*Ulysses*"), Russian (Leo N. Tolstoy, "*War and Peace*"; Mikhail A. Sholokhov, "*And Quiet Flows the Don*") and Ukrainian (Pavlo A. Zahrebelnyi "*Roksolana*") (see Table 1) ranging from 85.000 to 428.000 "types" or "expression units" (abbreviated as EUs, (Pustet, 2004; see Table 2). In parallel, two program files in Java and C ++ were treated in a similar way. Before the analysis, all punctuation symbols (such as points, commas, question and exclamation marks, dashes, etc. which have been suggested to be semantically negligible) as well as footnotes and notes were removed from the text files. The number of unique words in the texts representing natural languages, and the number of repetitions of each unique word (EU). These were determined with default delimiters such as {' ' (space), '\f', '\n', '\r', '\t', '\v'} were calculated with software based on the Matlab.

In contrast, the program languages were treated as follows: only the texts of comments were deleted, whereas all the marks of punctuation (such as point, comma, etc.) have been considered as EUs, since any part of the program (such as operators and arguments) is semantically significant. When Java and C ++ files were processed, all operators, variables and comment markers were separated by a 'space' symbol, and all the comments were deleted. As an example, our treatment examined the text of "*War and Peace*" and the program code written in Java. The quantitative data related to these texts are displayed in Tables 3a-3b.

## 3. Results

The data appearing in the second column of Tables 3a-3b (namely, the total number of appearances of EUs in the text) formed the random set of numbers. The validity of the Benford distribution of leading numbers was checked for this set. Frequencies of specific EUs, summarized in the third column, were checked from the point of view of the Zipf distribution, as shown in Figures 1a-7a. It was suggested that Zipf's law is valid, but not for the most common words (Tsonis et al., 1997a). For examination of this hypothesis, the following procedure was applied for all of the analyzed texts (written in human and "artificial" languages): first, the most common words were eliminated and the Zipfian approximation procedure was repeated, as displayed in Figures 1b-7b. The Benford approximation of leading digits to the actual one for the set, formed by total numbers of appearances of EUs in the text of "*Ulysses*", is presented in Figure 8. The quantitative parameters of the texts established with the Zipf and Benford approximations are summarized in Table 3. We draw the attention of the reader to a visual artifact arising from the graphs, depicted in Figures 1b-7b. It seems that the correlation of actual text data, obtained after excluding the most popular words with the Zipfian distribution is low (the "blue circles" and "red dashed" curves do not coincide satisfactorily). This visual artifact is due to the fact that the graphs displayed in Figures 1b-7b are restricted by $n = 110$, at their abscissa. The artifact almost disappears when the graphs are extended to $n = 300$, as shown in Figure 1a. The double logarithmic ($lnf(n)$ vs. $ln(n)$) plot of the same data, supplied in Figs.1c-7c and Figs. 1d-7d is much more representative.

## 4. Discussion

First of all, it should be concluded that both the distributions of EUs and leading numbers in total numbers of appearances of unique words in a text are markedly uneven. It is seen from the supplied correlation coefficients of the actual distribution of frequencies of EUs with the Zipf law that it essentially increased in the case where the first ten words were eliminated (see and compare data, illustrated by Figures 1-7). This was true for both human and artificial languages (namely Java and C++). When the most common words are excluded, the distribution of frequencies of EUs is definitely Zipf-like. The moduli of slopes of *lnf(n)* vs. *ln(n)* plots for all of studied human languages were restricted within $0.86 < \alpha < 1.03$.

The programs written in investigated artificial languages (C++ and Java) demonstrated Zipf-like behavior when the most common operators and arguments were excluded, the double logarithmic plot of *lnf(n)* vs. *ln(n)* was linear at least at the "tail" of distributions, as shown in Fig. 6d and Fig. 7d. However, the modulus of the slope at the "tail" of distributions was larger and far from the unity, inherent for texts written in human languages. The value of the modulus of the slope for software written with Java was established as *α* = 1.25, and it is very close to the value reported already by Zhang (Zhang, 2009). The software written in the C++ demonstrated more complicated behavior. The frequency of the most common operators (*n*<11) appearing in C++ is not described by the power law; for the ranks, restricted within $12 < n < 33$ the double logarithmical plot demonstrated the value of *α* close to unity, whereas the modulus of the slope established for the tail of the *lnf(n)* vs. *ln(n* dependence was as high as *α* = 2.39 (for *n* >33, as shown in Fig.7d). It is plausible to suggest that the Zipf-Mandelbrot law is more appropriate for describing data, arising from texts (software), written in artificial languages (Zhang, 2009). The reported findings support the idea that the hidden quantitative patterns of human and artificial languages are similar (Zhang, 2009).

As to the set of numbers formed by the total number of appearances of a certain EU in the text we recognize that the distribution of leading digits in it is pronouncedly uneven; however it strictly follows the Benford distribution only in "*Ulysses*" and Java software. For the other texts, the coefficients of correlation are relatively low, and it should be emphasized that the approximation procedure turned out to be insensitive to the elimination of the most common words, for all the studied human and artificial languages (software). For example, the elimination of the most popular words in the text of "*Harry Potter and the Chamber of Secrets*" changed the correlation coefficient for Benford's distribution from 0.970864 to 0.970883; this means that the correlation coefficient increased by 0.002%; for "*Ulysses*" the established change in the correlation coefficient was negative and its modulus was 0.0015%.

Let us check one more observation related to the Benford distribution. Since Benford's seminal paper, many investigations have shown that amalgamating data from different sources leads to Benford behavior (Benford, 1938; Miller & Nigrini, 2008; Bormashenko et al., 2016). Indeed, amalgamation of the Benford data extracted from all of the treated text written with human and artificial languages supplied a coefficient of correlation with the Benford law as high as 0.968. The reasonable question is: why is the correlation of the textual data with the Benford law not perfect? The answer is that the texts analyzed in the paper contained an enormous amount of unique EUs; thus the number "1" appearing as a leading digit was disproportionally large (see Table 2).

**Conclusions**

We conclude that texts written in natural (in other words, human) languages, represented in our study by English, Russian and Ukranian, and artificial ones, such as C++ and Java, are characterized by hidden quantitative patterns, which are apparently universal. The distribution of frequencies of words follows the Zipf law (Zipf, 1965); whereas the total numbers of a certain expression unit, appearing in the text generate a non-even distribution of leading digits, which is close to the logarithmical distribution, introduced by Newcomb and Benford (Newcomb, 1881; Benford, 1938). It is noteworthy that the studied human languages represent different families: English belongs to the Germanic family, while Russian and Ukranian represent the Slavic family of languages. Thus, we come to the conclusion that a co-occurrence of Zipf and Benford-like laws takes place for data sets generated by large texts, written in human and artificial languages. Zipf-like power-law regularities, appearing in large-scale software systems, prepared with Java and C++ were reported recently (Zhang, 2009). These findings support the idea that the Zipf-like behavior originates in the random group formation scheme, common for a broad diversity of statistical data (Baek et al., 2011).

After the elimination of the most popular words, the coefficients of correlation of quantitative data extracted from "*Ulysses*", "*Harry Potter and the Chamber of Secrets*" and "*War and Peace*" approximated with the Zipf law were as high as 0.999, 0.998 and 0.997 respectively. These findings strengthen the hypothesis that the classical Zipfian law, namely: $f(n) = \dfrac{C}{n^\alpha}$ (where $f$ is the frequency of an expression unit and $n$ is its Zipf rank) is valid for texts. Whereas other functions describing the distribution of frequencies of expression units (such as $f \sim \dfrac{\exp(-bn)}{n^\gamma}; \gamma = const; b = const$) were introduced and discussed (Baek et al., 2011; Cristelli et al., 2012). It is also noteworthy that the Zipf law works better for texts written with human languages than for software. The established values of the moduli of slopes *α* for artificial languages are markedly larger than those for human ones.

Is the co-occurrence of the Zipf and Benford laws in the explored textual data sets accidental? We suggest that the answer is negative, and that a close relationship between these distributions does exist. It was demonstrated recently that if a function, describing the dependence between two measurable quantities has a positive second derivative, then the frequencies *f* of decimal digits at the first place of numbers decrease for digits from 1 to 9 (Whyman et al., 2016b). And the Zipfian frequency ($f(n) = \dfrac{1}{n^{\alpha}}, \alpha > 0$) obviously fulfils this demand. The function describing this decrease is not necessarily logarithmical (namely given by Eq. 2). Actually our results demonstrate that at best, the distribution of leading digits is uneven, decreasing and may be defined as "Benford-like" (quasi-Benford), due to the not-too-high values of the correlation coefficients relating the actual data extracted from the texts to the Benford distribution, summarized in Table 2. The best correlation with the Benford distribution was demonstrated in programs written in Java, with a coefficient of correlation as high as 0.993. This begs the question: what is unique in software prepared with Java? We hope that further insights into the problem will supply the answer.

Table 1. The texts treated in the article.

| Text # | Author and title | Language |
| --- | --- | --- |
| 1 | Joanne K. Rowling, "*Harry Potter and the Chamber of Secrets*" | English |
| 2 | James Joyce, "*Ulysses*" | English |
| 3 | Leo N. Tolstoy, "*War and Peace*" | Russian |
| 4 | Mikhail A. Sholokhov, "*And Quiet Flows the Don*" | Russian |
| 5 | Pavlo A. Zahrebelnyi ,"*Roksolana*" | Ukrainian |
| 6 | Program file 83 kB | Java |
| 7 | Program file 148 kB | C++ |

Table 2. The quantitative parameters of the texts, as established with the Zipf and Benford approximations.

| Text # | Total words quantity | Unique words quantity | Correlation coefficient with the Zipf law | Correlation coefficient with the Zipf law in the case when the first ten words were eliminated | Correlation coefficient with the Benford law and percent of leading "1" |
| --- | --- | --- | --- | --- | --- |
| 1 | 85159 | 7766 | 0.926 | 0.998 | 0.971 (55%) |
| 2 | 302001 | 30104 | 0.967 | 0.999 | 0.989 (50%) |
| 3 | 230085 | 34136 | 0.964 | 0.997 | 0.960 (62%) |
| 4 | 428015 | 73471 | 0.940 | 0.995 | 0.963 (61%) |
| 5 | 228664 | 40660 | 0.968 | 0.992 | 0.961 (62%) |
| 6 | 21364 | 863 | 0.884 | 0.984 | 0.993 (40%) |
| 7 | 35012 | 529 | 0.899 | 0.982 | 0.918 (29%) |

Table 3a. The quantitative analysis of "*War and Peace*".

| Expression Unit | Number of appearances of EU in text | Frequency | |
|---|---|---|---|
| | | Actual | As predicted by Zipf's law |
| 'И' | 10622 | 0.0462 | 0.0908 |
| 'В' | 5286 | 0.0230 | 0.0454 |
| 'НЕ' | 4415 | 0.0192 | 0.0303 |
| 'ЧТО' | 3946 | 0.0172 | 0.0227 |
| 'ОН' | 3823 | 0.0166 | 0.0182 |
| 'НА' | 3347 | 0.0145 | 0.0151 |
| 'С' | 3100 | 0.0135 | 0.0130 |
| 'КАК' | 2147 | 0.0093 | 0.0113 |
| 'ЕГО' | 1938 | 0.0084 | 0.0101 |
| 'Я' | 1934 | 0.0084 | 0.0091 |

Table 3b. The quantitative analysis of the program code written in Java.

| Expression Unit | Number of appearances of EU in text | Frequency | |
|---|---|---|---|
| | | Actual | As predicted by Zipf's law |
| ';' | 1607 | 0.0752 | 0.1363 |
| '(' | 1497 | 0.0701 | 0.0681 |
| ')' | 1497 | 0.0701 | 0.0454 |
| '.' | 1315 | 0.0616 | 0.0341 |
| ',' | 1160 | 0.0543 | 0.0273 |
| '=' | 906 | 0.0424 | 0.0227 |
| 'INT' | 578 | 0.0271 | 0.0195 |
| '{' | 530 | 0.0248 | 0.0170 |
| '}' | 530 | 0.0248 | 0.0151 |
| 'X' | 355 | 0.0166 | 0.0136 |

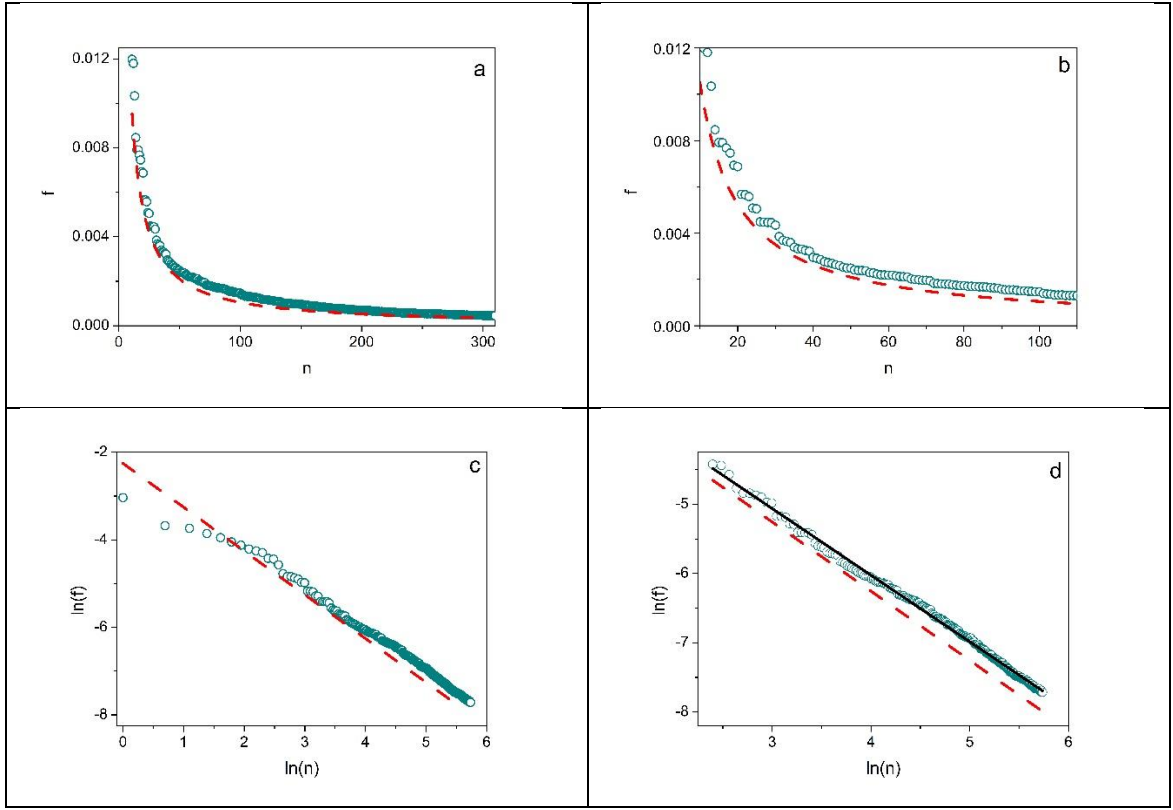

Fig 1. The text of "*Harry Potter and the Chamber of Secrets*" treated according to Zipf's approximation. The red dashed line in all of the graphs represents the Zipf curve ($α=1$, see Eq. 1); the open circles represent the actual distribution of EUs. a) The first three hundred points of the dependence *f(n)* are presented, where *f* is the frequency, and *n* is the Zipf rank of a word (EU). The coefficient of correlation as calculated with the least squares method was 0.926. b) The graph starts from $n = 11$ and expands to $n = 110$. The coefficient of correlation as calculated with the least squares method was 0.998. c) The dependence *ln(f(n))* vs. *ln(n)* is presented. The graph starts from $n = 1$ and expands to $n = 310$. d) The dependence *ln(f(n))* vs. *ln(n)* is presented. The graph starts from $n = 11$ and expands to $n = 310$. The solid line represents the best linear fit for the actual distribution of EUs and the modulus of its slope is $α = 0.96$.

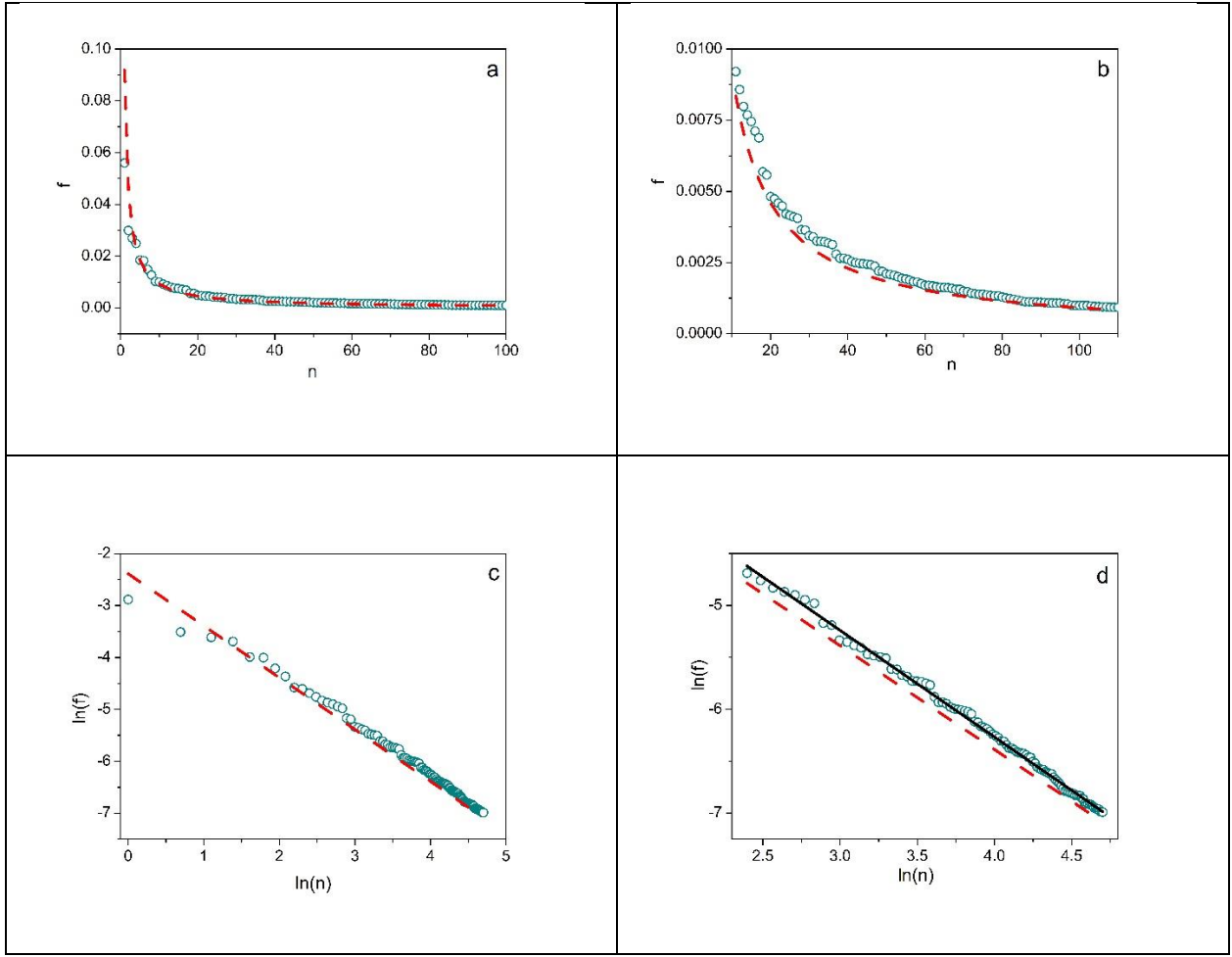

Fig 2. The text of "*Ulysses*" treated according to Zipf's approximation. The red dashed line represents in all of the graphs the Zipf curve ($α=1$, see Eq.1); the open circles represent the actual distribution of EUs. a) The first hundred points of the dependence *f(n)* are presented, where *f* is the frequency, and *n* is the Zipf rank of a word (EU). The coefficient of correlation as calculated with the least squares method was 0.967. b) The graph starts from $n = 11$ and expands to $n = 110$. The coefficient of correlation as calculated with the least squares method was 0.999. c) The dependence *ln(f(n))* vs. *ln(n)* is presented. The graph starts from $n = 1$ and expands to $n = 110$. d) The dependence *ln(f(n))* vs. *ln(n)* is presented. The graph starts from $n = 11$ and expands to $n = 110$. The solid line represents the best linear fit for the actual distribution of EUs and the modulus of its slope is $α = 1.03$.

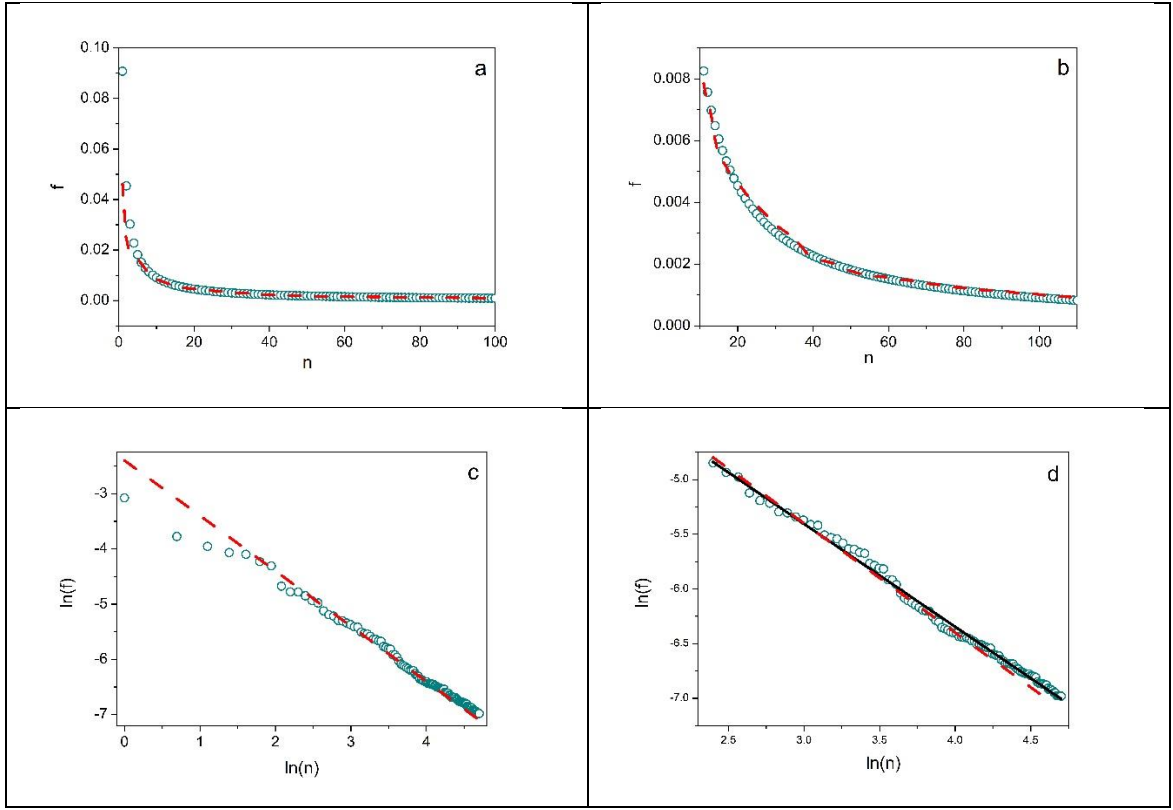

Fig 3. The text of "*War and Peace*" treated according to Zipf's procedure. The red dashed line represents the Zipf curve ($\alpha=1$); the open circles represent the actual distribution of EUs. a) The first hundred points of the dependence *f(n)* are presented, where *f* is the frequency, and *n* is the Zipf rank of a word (EU). The coefficient of correlation as calculated with the least squares method was 0.964. b) The graph starts from $n = 11$ and expands to $n = 110$. The coefficient of correlation as calculated with the least squares method was 0.997. c) The dependence *ln(f(n))* vs. *ln(n)* is presented. The graph starts from $n = 1$ and expands to $n = 110$. d) The dependence *ln(f(n))* vs. *ln(n)* is presented. The graph starts from $n = 11$ and expands to $n = 110$. The solid line represents the best linear fit for the actual distribution of EUs, where the modulus of its slope is $\alpha = 0.94$.

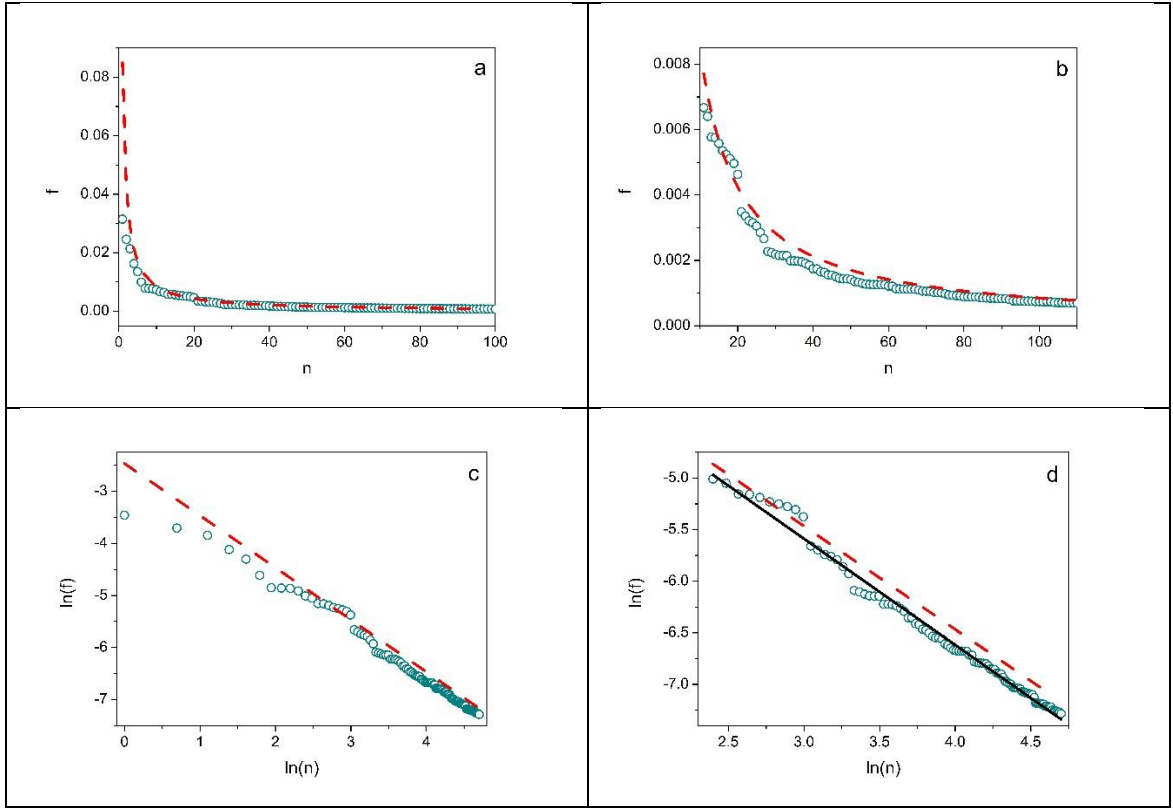

Fig 4. The text of "*And Quiet Flows the Don*" treated according to Zipf's approximation. The red dashed line represents the Zipf curve ($\alpha=1$) ; the open circles represent the actual distribution of EUs. a) The first hundred points of the dependence *f(n)* are presented, where *f* is the frequency, and *n* is the Zipf rank of a word (EU).The coefficient of correlation as calculated with the least squares method was 0.940. b) The graph starts from $n = 11$ and expands to $n = 110$. The coefficient of correlation as calculated with the least squares method was 0.995. c) The dependence *ln(f(n))* vs. *ln(n)* is presented. The graph starts from $n = 1$ and expands to $n = 110$. d) The dependence *ln(f(n))* vs. *ln(n)* is presented. The graph starts from $n = 11$ and expands to $n = 110$. The solid line represents the best linear fit for the actual distribution of EUs, where the modulus of its slope is $\alpha = 1.03$.

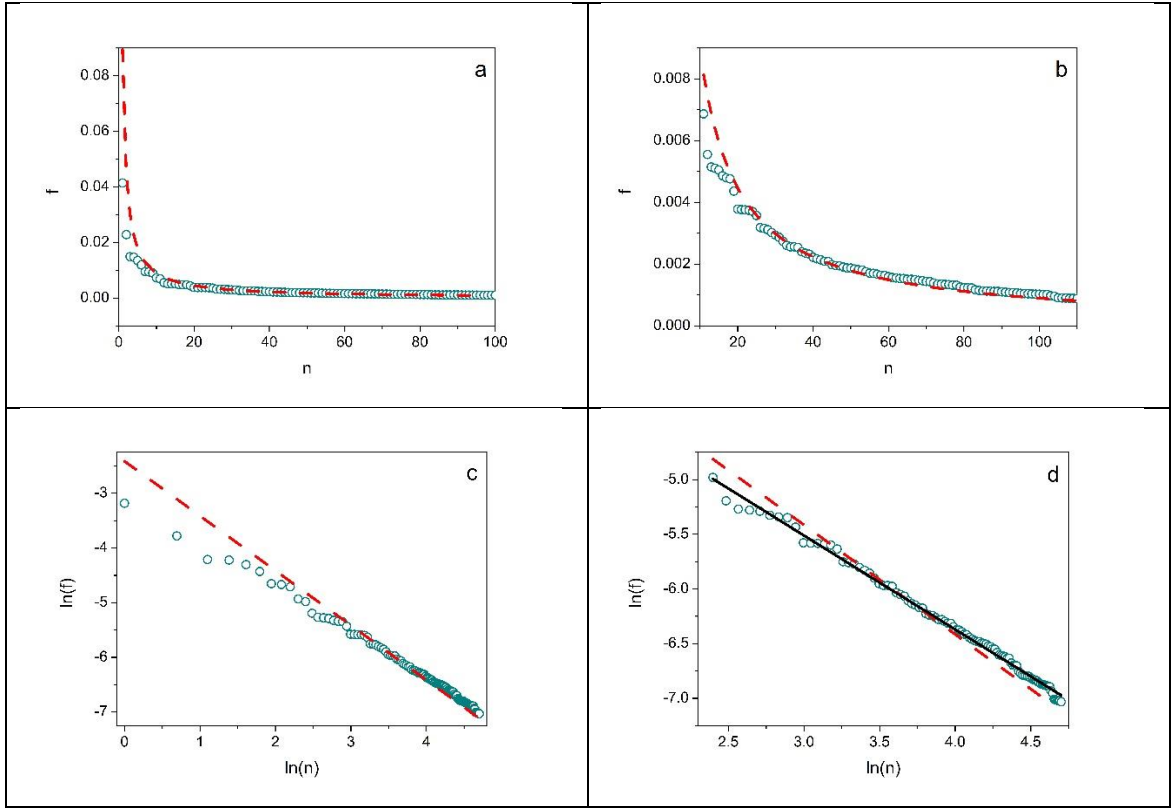

Fig 5. The text of "Roksolana" treated according to Zipf's approximation. The red dashed line represents the Zipf curve ($\alpha=1$); the open circles represent the actual distribution of EUs. a) The first hundred points of the dependence $f(n)$ are presented, where $f$ is the frequency, and $n$ is the Zipf rank of a word (EU). The coefficient of correlation as calculated with the least squares method was 0.968. b) The graph starts from $n = 11$ and expands to $n = 110$. The coefficient of correlation as calculated with the least squares method was 0.992. c) The dependence $ln(f(n))$ vs. $ln(n)$ is presented. The graph starts from $n = 1$ and expands to $n = 110$. d) The dependence $ln(f(n))$ vs. $ln(n)$ is presented. The graph starts from $n = 11$ and expands to $n = 110$. The solid line represents linear fit for the actual distribution of EUs, where the modulus of its slope is $\alpha = 0.86$.

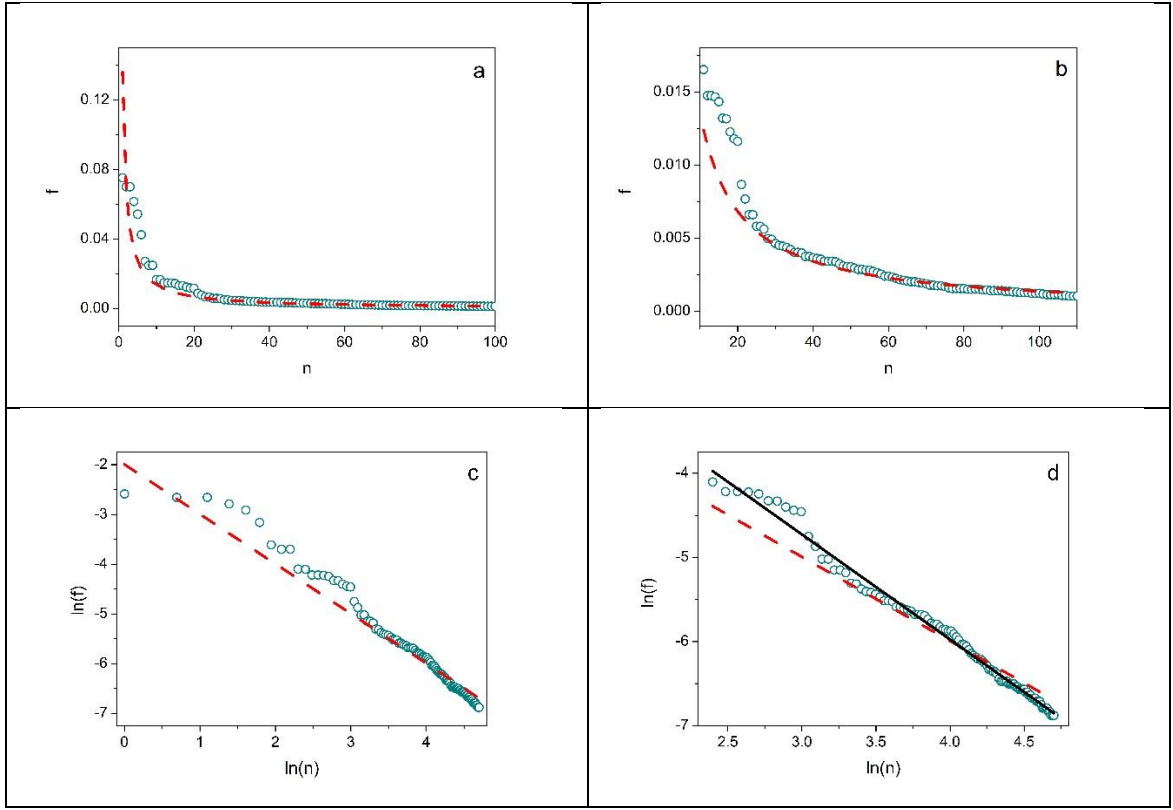

Fig 6. The text of the program in Java treated according to Zipf's approximation. The red dashed line represents the Zipf curve (*α*=1); the open circles represent the actual distribution of EUs. a) The first hundred points of the dependence *f(n)* are presented, where *f* is the frequency, and *n* is the Zipf rank of a word (EU). The coefficient of correlation as calculated with the least squares method was 0.884. b) The graph starts from *n* = 11 and expands to *n* = 110. The coefficient of correlation as calculated with the least squares method was 0.984. c) The dependence *ln(f(n))* vs. *ln(n)* is presented. The graph starts from *n* = 1 and expands to *n* = 110. d) The dependence *ln(f(n))* vs. *ln(n)* is presented. The graph starts from *n* = 11 and expands to *n* = 110. The solid line represents the best linear fit for the actual distribution of EUs, where the modulus of its slope is *α* = 1.25.

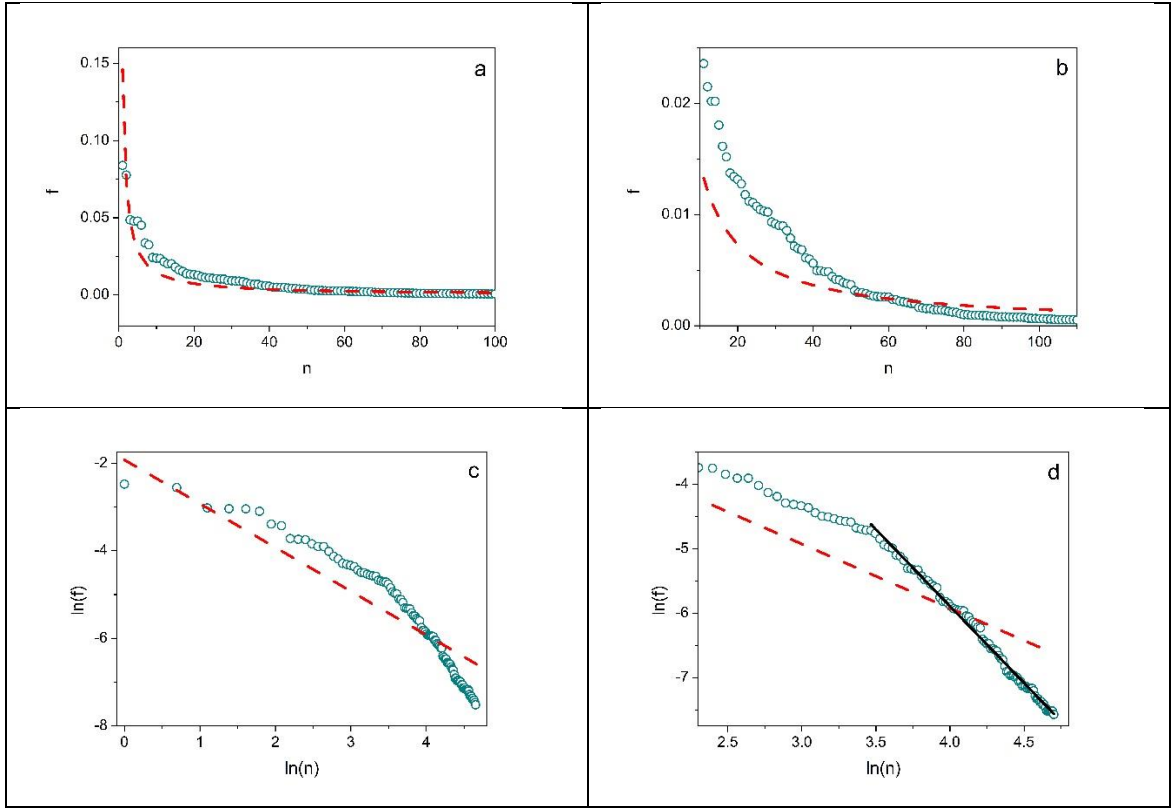

Fig 7. The text of the program in C++ treated according to Zipf's approximation. The dashed line represents the Zipf curve ($\alpha=1$); the open circles represent the actual distribution of EUs. a) The first hundred points of the dependence *f(n)* are presented, where *f* is the frequency, and *n* is the Zipf rank of a word (EU). The coefficient of correlation as calculated with the least squares method was 0.899. b) The graph starts from $n = 11$ and expands to $n = 110$. The coefficient of correlation as calculated with the least squares method was 0.982. c) The dependence *ln(f(n))* vs. *ln(n)* is presented. The graph starts from $n = 1$ and expands to $n = 110$. d) The dependence *ln(f(n))* vs. *ln(n)* is presented. The graph starts from $n = 11$ and expands to $n = 110$. The solid line represents the best linear fit for the actual distribution of EUs (for *n>33*) where the modulus of its slope is $\alpha = 2.39$.

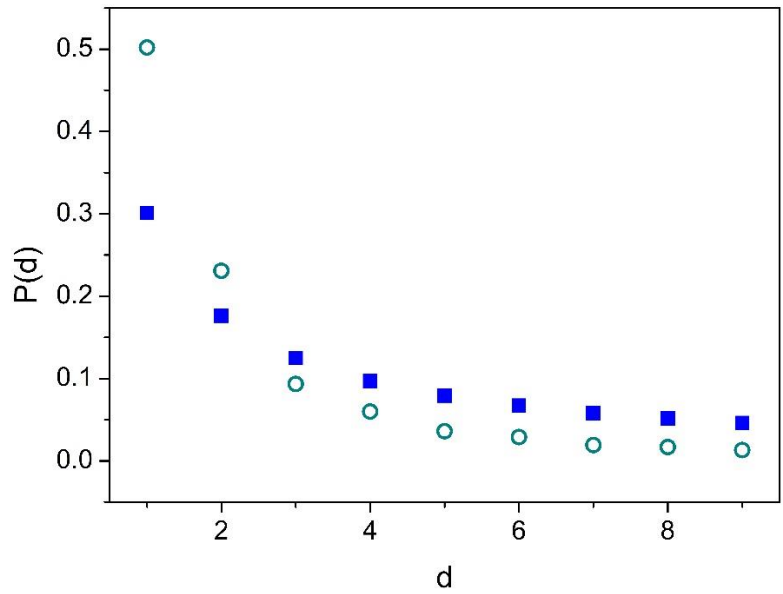

Fig. 8. The actual distribution of leading digits for the set formed by total numbers of appearances of EUs in the text of "*Ulysses*" is shown, when compared to Benford's Law. The squares represent logarithmical Benford's distribution; the open circles represent the actual distribution of first (leading) digits.